\documentclass[runningheads]{llncs}
\usepackage{graphicx}
\usepackage{enumerate}
\usepackage{paralist}
\usepackage{url}
\usepackage{xcolor,colortbl}
\usepackage{hyperref}
\usepackage{comment}
\usepackage{multirow}
\usepackage{latexsym}
\usepackage[inline, shortlabels]{enumitem}
\usepackage{paralist}
\usepackage{multirow}
\usepackage{xcolor,colortbl}
\begin{document}
%

\title{Selection of pseudo-annotated data for adverse drug reaction classification across drug groups}
\titlerunning{Pseudo-annotated data for ADR classification across drug groups}

\author{Ilseyar Alimova\inst{1} \and Elena Tutubalina\inst{1,2,3}}
\institute{Kazan (Volga Region) Federal University, Kazan, Russia \and
HSE University, Russia \and Sber AI, Russia}

%
%
%
\maketitle              
\begin{abstract}
Automatic monitoring of adverse drug events (ADEs) or reactions (ADRs) is currently receiving significant attention from the biomedical community. In recent years, user-generated data on social media has become a valuable resource for this task. Neural models have achieved impressive performance on automatic text classification for ADR detection. Yet, training and evaluation of these methods are carried out on user-generated texts about a targeted drug. In this paper, we assess the robustness of state-of-the-art neural architectures across different drug groups. We investigate several strategies to use pseudo-labeled data in addition to a manually annotated train set. Out-of-dataset experiments diagnose the bottleneck of supervised models in terms of breakdown performance, while additional pseudo-labeled data improves overall results regardless of the text selection strategy.


\end{abstract}
\section{Introduction}
Pharmacovigilance from social media data that focuses on discovering adverse drug effects (ADEs) from user-generated texts (UGTs). ADEs\footnote{The terms ADEs and adverse drug reactions (ADRs) are often used interchangeably.} are unwanted negative effects of a drug, in other words, harmful and undesired reactions due to its intake. 

In recent years, researchers have increasingly applied neural networks, including Bidirectional Encoder Representations from Transformers (BERT) \cite{devlin2018bert}, to various biomedical tasks, including as text-level or entity-level ADR classification of user-generated texts \cite{klein2020overview,magge2021overview,tutubalina2018exploring,alimova2020multiple}. The text-level ADR classification task aims to detect whether a given short text contains a mention of an adverse drug effect. The entity-level task focuses on the classification of a biomedical entity or a phrase within a text. The first type of classification is needed to filter irrelevant texts from a data collection. In the second case, classification models process results of named entity recognition (NER) tools. However, most recent studies mostly share the same limitations regarding their training strategy: classification systems rely only on existing manually annotated training data for supervised machine learning.  
These annotated corpora include texts about a small number of drugs (at best about dozens) while there are over 20,000 prescription drug products approved for marketing\footnote{https://www.fda.gov/about-fda/fda-basics/fact-sheet-fda-glance}. Moreover, the model performance is frequently evaluated under the implicit hypothesis that the training data (source) and the test data (target) come from the same underlying distribution (i.e., both sets include drugs from a particular Anatomical Therapeutic Chemical (ATC) group). This hypothesis can cause overestimated results on ADR classification since the same drug classes shared similar patterns of ADR presence \cite{wu2019study}.

\begin{figure}[t]
    \centering
    \includegraphics[width=0.9\textwidth]{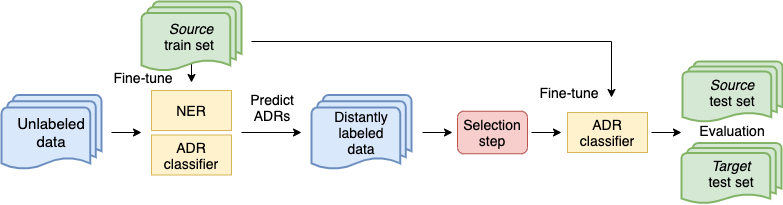}
    \caption{An overview of our pipeline.}
    \label{fig:framework}
\end{figure}

In this paper, we take the task a step further from existing research by exploring how well a BERT-based classification model trained on texts about drugs from one ATC group (\textit{source}) performs on texts about other drugs from the second ATC group (\textit{target}). First, we train a base model on some amount of \textit{source} labeled data and use this model to pseudo-annotate unlabeled data. Second, the original labeled data is augmented with the pseudo-labeled data and used to train a new model. Third, both models are evaluated on source data (\textit{in-dataset}) and target data (\textit{out-of-dataset}) (Fig. \ref{fig:framework}).
In particular, we focus on the automatic expansion of training data.
We explore several strategies to select a subset to train our ADR classification model for target data.  In this work, we seek to answer the following research questions:
\begin{enumerate}[start=1,label={\bfseries RQ\arabic*:}]
\item Do in-dataset evaluation with training and testing on each benchmark separately lead to a significant overestimation of performance? 
\item How can we utilize raw reviews without manual annotations to improve model performance on a particular drug group?
\end{enumerate}


\section{Related Work}\label{sec:related}
ADR classification is usually formulated as a classification or ranking problem. A number of supervised neural models have been proposed; please see recent state-of-the-art models in Social Media Mining for Health shared tasks~\cite{klein2020overview,magge2021overview}. Further, we describe several studies that leveraged different types of unlabeled data to deal with data bias or a small number of labeled samples. 

The work that is the closest to ours and considers unlabeled data is \cite{lee2017adverse}. The authors proposed utilized semi-supervised convolutional neural network models for ADR classification in tweets. The method works in two phases: (1) unsupervised phrase embedding learning on unlabeled data, and (2) integrating the learned embeddings into the supervised training that uses labeled data. 
The authors used several types of unlabeled data: (i) collection of random tweets; (ii) corpus of health-related texts; (iii) tweets with drug names; (iv) tweets that mention a health condition. Health-related texts include sentences from the clinical medicine category Wikipedia pages, PubMed articles, and UMLS Medical Concept Definitions. These types of unlabeled data were used to learn phrase embeddings for the semi-supervised classification. The experiments show that the model trained on the tweets with health conditions outperformed in average models trained on other unlabeled datasets by 3.3\% of F-score. This model also outperformed trained only on annotated dataset model by 8.9\% of F-score. 
Gupta et al. present a semi-supervised Bi-directional LSTM based model for ADR mention extraction from tweets \cite{gupta2018semi}. 
The large corpus of unlabeled tweets was collected for unsupervised learning. The semi-supervised approach improved the F1-score metric by 2.2\%. \cite{gupta2018co} proposed a semi-supervised co-training based learning method to tackle the problem of labeled data scarcity for adverse drug reaction mention extraction task. The methods were evaluated on two corpora of tweets. The results show that increasing the dataset leads to improving results. Thus, the model co-trained on 100K tweets outperformed baselines on F-Score by 6.52\% and 4.8\% on Twitter ADR and TwiMed corpora, respectively.
Perez et al. investigated the semi-supervised approach for detecting ADR mentions on Spanish and Swedish clinical corpora \cite{perez2017semi}. The authors applied unlabeled data of EHR texts to obtain Brown trees and semantic space cluster features. The features were enhanced to maximum probability, CRF, Perceptron, and SVM classifiers. 
The results showed that the semi-supervised approaches significantly improved standard supervised techniques for both languages on average by 12.26\% and 4.56\% of F-score for Spanish and Swedish languages, respectively.

Our work differs from the studies discussed above in the following important aspects. First, we annotate unlabeled data with noise labels and use it to train for a classification model instead of language pre-training. Second, we utilize state-of-the-art BERT-based models. Finally, we evaluate models across different drug groups with in-dataset and out-of-dataset setup.

\section{Datasets and Models}\label{sec:data}

We perform all experiments in four steps (Fig. \ref{fig:framework}). First, we train two models on source data: (i) a classifier for ADR identification, (i) a NER model for detection of biomedical disease-related entities. Second, we apply both models to extract medical entities from an unlabeled corpus and classify these entities as ADRs. Third, we select a set of texts with ADR mentions. Finally, we train the ADR classifier from Step 1 on the pseudo-annotated data and evaluate this model on the target corpus. Further, we describe two manually annotated datasets, a raw corpus of reviews, and neural models.



\subsubsection{CADEC} CSIRO Adverse Drug Event Corpus (CADEC) \cite{karimi2015cadec} consists of annotated user reviews from \url{https://www.askapatient.com/} about 12 drugs, divided into two groups: 
\begin{inparaenum}
\item Diclofenac; 
\item Lipitor. 
\end{inparaenum}

Diclofenac is linked with the ATC group \textit{Nervous system} (N). 
Lipitor is included into three ATC main groups: \textit{Sensory organs} (S), \textit{Musculoskeletal system} (M), \textit{Dermatologicals} (D). 
The corpus contains 6,320 entities, 5,770 of them marked as `ADR'. 
Following \cite{alimova2017automated,li2020exploiting,rakhsha2021detecting}, we group non-ADR types (diseases, symptoms and findings) in a single opposite class for binary classification.

\subsubsection{PsyTAR}

(Psychiatric Treatment Adverse Reactions (PsyTAR) corpus \cite{zolnoori2019systematic} is the first open-source corpus of user-generated posts about psychiatric drugs taken from \url{https://www.askapatient.com/}. This dataset includes 887 posts about four psychiatric medications of two classes: 
\begin{inparaenum}
    \item Zoloft and Lexapro;
    \item Effexor and Cymbalta.
\end{inparaenum}

Zoloft,  Lexapro, Effexor, and Cymbalta are included in the ATC group \textit{Nervous system} (N). 
All posts were annotated manually for 4 types of entities:
\begin{inparaenum}[(i)]
    \item adverse drug reactions (ADR);
    \item withdrawal symptoms (WD);
    \item drug indications (DI);
    \item sign/symptoms/illness (SSI).
\end{inparaenum}
The total number of entities is 7,415. We joined WD, DI and SSI entities in a single class. 



\subsubsection{Unlabeled Data}


We collect 113,836 reviews about 1,593 drugs from \url{https://www.askapatient.com/}. There are 35,712 reviews with a rating 1. The total number of reviews about drugs from the CADEC corpus is 173 and from the PsyTAR corpus is 6,590. We create several subsets for our experiments:
\begin{enumerate}
    \item The full set of reviews. This set is abbreviated as AskaPatient$_{full}$;
    \item The set of reviews about target drugs (AskaPatient$_{target}$);
    \item The set of reviews with the lowest rating (AskaPatient$_{1}$). 
    Each review is associated with an overall rating, which is a numeric score between 1 (dissatisfied) and 5 (very satisfied). The lowest rating indicates that a user would not recommend taking this medicine. 
\end{enumerate}



\subsubsection{Models}
In particular, we utilize two models: 
\begin{enumerate}
    \item LSTM-based Interactive Attention Network (IAN) \cite{ma2017interactive};
    \item BERT \cite{devlin2018bert,lee2020biobert}.
\end{enumerate}
\cite{alimova2019entity} showed the superiority of IAN over other neural models on
four of the five corpora for entity-level ADR text classification. 
We used 15 epochs to train IAN on each dataset, the batch size of 128, the number of hidden units for LSTM layer 300, the learning rate of 0.01, L2 regularization of 0.001, dropout 0.5. We applied the implementation of the model from \url{https://github.com/songyouwei/ABSA-PyTorch}. We trained BioBERT model \cite{lee2020biobert} for 15 epochs. 

\section{Experiments}

\begin{table}[!t]
\centering
\caption{Classification results of IAN and BERT evaluated on CADEC.}\label{tab:results1}
\setlength{\tabcolsep}{4pt}
\begin{tabular}{|l|l|ccc|ccc|}
\hline 
\textbf{Train set} & \textbf{Model} & \multicolumn{3}{c|}{\textbf{ADR-class}} & \multicolumn{3}{c|}{\textbf{Macro-averaged}} \\
\cline{3-8}
& & P & R & F & P & R & F \\
\hline
\multicolumn{8}{|c|}{In-dataset performance} \\ \hline
CADEC  & IAN & .966 & .972 & .969 & .832 & .805 & .815 \\ 
CADEC & BERT & .947 & .983 & .965 & .824 & .702 & .746\\
\hline
\multicolumn{8}{|c|}{Out-of-dataset performance} \\ \hline
PsyTAR & IAN & .898 & .885 & .891 & .615 & .624 & .619 \\
PsyTAR & BERT & .909 & .914 & .911 & .673 & .669 & .671\\
\hline
PsyTAR + AskaPatient$_{full}$& IAN &  .903 & .895 & .899 & .636 & .642 & .638\\
PsyTAR + AskaPatient$_{1}$& IAN & .896 & .951 & .922 & .693 & .626 & .647\\
PsyTAR + AskaPatient$_{target}$ & IAN &.950 & .968 & .959 & .861 & .823 & .841\\
\hline
PsyTAR + AskaPatient$_{full}$ & BERT &  .912 & .918 & .915 & .681 & .675 & .677\\
PsyTAR + AskaPatient$_{1}$ & BERT & .959 & .839 & .895 & .695 & .806 & .724\\
PsyTAR + AskaPatient$_{target}$ & BERT & .952 & .929 & .941 & .782 & .817 & .798 \\
\hline
\end{tabular}
\end{table}

\begin{table}[!t]
\centering
\caption{Classification results of IAN and BERT evaluated on PsyTAR.}\label{tab:results2}
\setlength{\tabcolsep}{4pt}
\begin{tabular}{|l|l|ccc|ccc|}
\hline 
\textbf{Train set} & \textbf{Model} & \multicolumn{3}{c|}{\textbf{ADR-class}} & \multicolumn{3}{c|}{\textbf{Macro-averaged}} \\
\hline
\multicolumn{8}{|c|}{In-dataset performance} \\ \hline
PsyTAR & IAN & .902 & .913 & .909 & .868 & .868 & .868 \\ 
PsyTAR & BERT & .904 & .917 & .910 & .884 & .882 & .881\\
\hline
\multicolumn{8}{|c|}{Out-of-dataset performance} \\ \hline
CADEC  & IAN &  .685 & .982 & .807 & .765 & .582 & .553\\
CADEC & BERT &  .711 & .984 & .825 & .808 & .629 & .623\\
\hline
CADEC + AskaPatient$_{full}$ & IAN & .693 & .836 & .751 & .591 & .577 & .563\\
CADEC + AskaPatient$_{1}$ & IAN & .766 & .931 & .841 & .784 & .703 & .721\\
CADEC + AskaPatient$_{target}$ &IAN &  .695 & .964 & .807 & .739 & .598 & .582  \\

\hline
CADEC + AskaPatient$_{full}$ & BERT &.718 & .986 & .831 & .810 & .627 & .630\\
CADEC + AskaPatient$_{1}$ & BERT &.764 & .973 & .856 & .834 & .714 & .731\\
CADEC + AskaPatient$_{target}$ & BERT & .681 & .978 & .803 & .746 & .573 & .633\\
\hline
\end{tabular}
\end{table}

All models were evaluated by 5-fold cross-validation. We computed averaged recall (R), precision (P), and $F_1$-measures (F) for ADR and non-ADR classes separately and then macro-average of these values for both classes.  The average F-score results for IAN and BERT models are presented in Tables \ref{tab:results1} and \ref{tab:results2}.

To answer \textbf{RQ1}, we compare in-dataset and out-of-dataset results in Tables \ref{tab:results1} and \ref{tab:results2}. The in-dataset performance of both models is significantly higher than out-of-dataset results. In particular, BERT trained on PsyTAR and CADEC achieved macro-averaged F$_1$ of 0.881 and 0.623 on the PsyTAR sets, respectively. This indicates the impact of different contexts and entity mentions on cross-dataset performance. 

To answer \textbf{RQ2}, we compare models trained on (i) manually annotated data and (ii) extended corpora with additional pseudo-annotated data. The results show that in cross-dataset experiments training the model on additional data improves results regardless of the selected noisy data in comparison to models trained only on labeled data. IAN and BERT trained on PsyTAR + Askapatient$_{target}$ showed the highest results on the CADEC corpus (0.841 and 0.798 F-score, respectively). These models outperformed models trained only on the CADEC corpus by 0.26 of F-score. On the PsyTAR corpus, the models trained on the CADEC + Askapatient$_1$ achieved the best results among models trained on additional noisy data (0.72). However, in this case, the F-score of the model trained on PsyTAR corpus is higher than the results of models trained on CADEC corpus and additional noisy data. Such a difference in results is due to reviews with a rating of 1 include 1,307 reviews about drugs from the PsyTAR corpus and do not contain reviews about drugs from the CADEC corpus.

Among the models additionally trained on noisy data the models trained on the full noisy corpus gave the smallest improvement in F-score metric on both corpora in comparison to models trained on annotated data. For the IAN model, the increase of F-score was 0.19 and 0.07, for the BERT model the improvement was 0.06 and 0.07 on the CADEC and PsyTAR cases, respectively.

Considering results in terms of `ADR' class metrics the same models achieved the best F-score metrics: trained on PsyTAR + AskaPatient$_{target}$ (IAN - 0.959 and BERT - 0.941) for CADEC corpus and trained on CADEC + AskaPatient$_1$ (IAN - 0.84 and BERT - 0.856) for PsyTAR corpus. However, in this case, none of the models outperformed the results of the model evaluated within a single corpus. Moreover, several models did not improve the F-score metrics in comparison to the same models trained only on annotated corpus: IAN trained on CADEC + AskaPatient$_{full}$ (0.75), BERT trained on PsyTAR + AskaPatient$_1$ (0.895) and BERT trained on CADEC + AskaPatient$_{target}$ (0.803).

\section{Conclusion} \label{sec:conclusion}

In this paper, we studied the task of discovering the presence of adverse drug effects in user reviews about anti-inflammatory and psychiatric drugs. We perform an extensive evaluation of LSTM-based and BERT-based models on two datasets in the cross-dataset setup. Our evaluation shows the great divergence in performance between splits of two corpora: BERT-based models trained on one corpus and evaluated on another show a substantial decrease in performance (-7.5\% and -25.8\% macro-averaged F$_1$ for the CADEC and PsyTAR sets, respectively) compared to in-dataset models. Using in-dataset models, we automatically annotate raw user posts and extend a train set with texts about a wide range of drugs. Retraining of models shows the increase in performance (e.g., up to +12.7\% macro-averaged F$_1$ for CADEC) compared to out-of-dataset models. We foresee three directions for future work. First, promising research directions include adaptive self-training and meta-learning techniques for training neural models with few labels. 
Second, future research may focus on advanced labeled data acquisition strategies such as uncertainty-based methods. Third, knowledge transfer from the clinical domain to the social media domain across different drug groups remains to be explored.

\subsubsection{Acknowledgments}
This work was supported by the Russian Science Foundation grant \# 18-11-00284.

%
%
%
%
\bibliographystyle{splncs04}
\bibliography{references}
\end{document}